\documentclass[10pt,twocolumn,letterpaper]{article}

\usepackage[pagenumbers]{cvpr} 
\usepackage[accsupp]{axessibility} 
\newcommand\blfootnote[1]{
  \begingroup
  \renewcommand\thefootnote{}\footnote{#1}
  \addtocounter{footnote}{-1}
  \endgroup
}

\usepackage[dvipsnames]{xcolor}
\usepackage{multirow}
\usepackage{algorithm}
\usepackage[noend]{algpseudocode}
\usepackage{mathtools}

\usepackage{amsmath,amsfonts,bm}

\def\eqref#1{equation~\ref{#1}}

\def\1{\bm{1}}

\def\vtheta{{\bm{\theta}}}

\def\ve{{\bm{e}}}

\def\vx{{\bm{x}}}
\def\vy{{\bm{y}}}

\DeclareMathAlphabet{\mathsfit}{\encodingdefault}{\sfdefault}{m}{sl}
\SetMathAlphabet{\mathsfit}{bold}{\encodingdefault}{\sfdefault}{bx}{n}

\def\gD{{\mathcal{D}}}

\def\gL{{\mathcal{L}}}

\DeclareMathOperator*{\argmin}{arg\,min}

\definecolor{cvprblue}{rgb}{0.21,0.49,0.74}
\usepackage[pagebackref,breaklinks,colorlinks,citecolor=cvprblue]{hyperref}
\def\paperID{0000} 
\def\confName{CVPR}
\def\confYear{2024}

\title{LAN: Learning to Adapt Noise for Image Denoising}

\author{Changjin Kim\\
Dept. of Computer Science\\
Hanyang University\\
{\tt\small chjinny@hanyang.ac.kr}
\and
Tae Hyun Kim\\
Dept. of Computer Science\\
Hanyang University\\
{\tt\small taehyunkim@hanyang.ac.kr}
\and
Sungyong Baik$^\dag$\\
Dept. of Data Science\\
Hanyang University\\
{\tt\small dsybaik@hanyang.ac.kr}
}
\begin{document}
\maketitle
\blfootnote{ $^\dag$ Corresponding author.}
\begin{abstract}
Removing noise from images, a.k.a image denoising, can be a very challenging task since the type and amount of noise can greatly vary for each image due to many factors including a camera model and capturing environments. 
While there have been striking improvements in image denoising with the emergence of advanced deep learning architectures and real-world datasets, recent denoising networks struggle to maintain performance on images with noise that has not been seen during training.
One typical approach to address the challenge would be to adapt a denoising network to new noise distribution.
Instead, in this work, we shift our focus to adapting the input noise itself, rather than adapting a network.
Thus, we keep a pretrained network frozen, and adapt an input noise to capture the fine-grained deviations.
As such, we propose a new denoising algorithm, dubbed Learning-to-Adapt-Noise (LAN), where a learnable noise offset is directly added to a given noisy image to bring a given input noise closer towards the noise distribution a denoising network is trained to handle.
Consequently, the proposed framework exhibits performance improvement on images with unseen noise, displaying the potential of the proposed research direction.  
The code is available at \href{https://github.com/chjinny/LAN}{https://github.com/chjinny/LAN}.
\end{abstract}
\vspace{-1.0em}
\section{Introduction}
\label{sec:intro}
Noise, an unwanted byproduct during image processing, can cause severe degradation not only in image quality but also in high-level computer vision tasks.
As such, noise has been a target to eliminate in the field of image denoising.
One of the main challenges in image denoising is how to distinguish a noise from an original source without knowing the noise distribution a priori.

To tackle such a challenging problem, a myriad of learning-based models have been proposed to learn to remove noise that follows known distributions.
To do so, models are trained on datasets composed of pairs of clean images and corresponding noisy images that are synthesized by adding noise to clean images.
Under such formulation, learning-based models have prominently marked progress in the performance of image denoising, ever since the emergence of convolutional neural networks (CNN) and its advanced architectures designed for image denoising~\cite{dncnn,hinet,anwar2019real,mirnet}.

\begin{figure}[t]
    \centering
\includegraphics[width=0.42\textwidth]{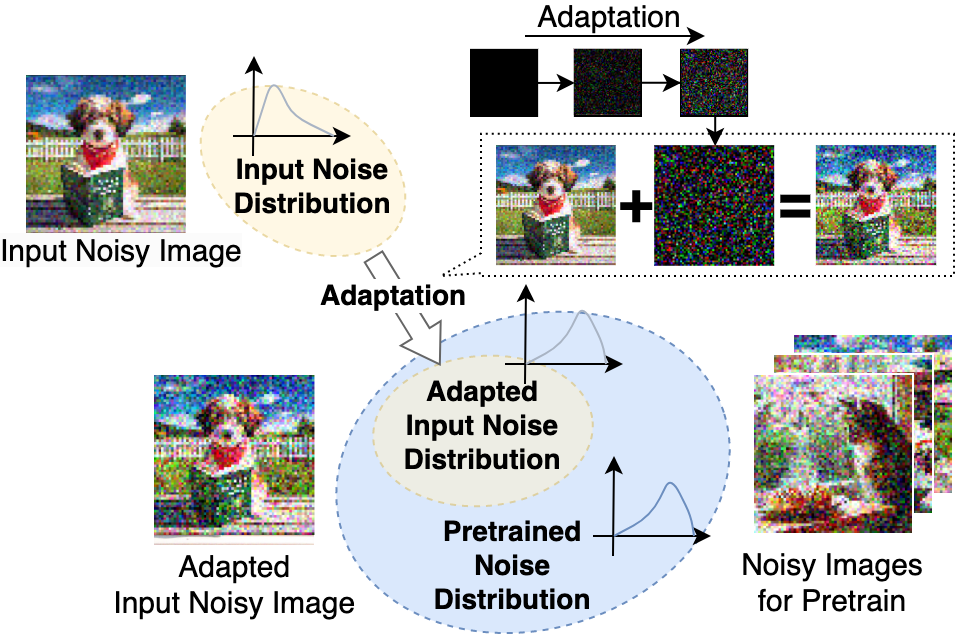}
    \vspace{-1.0em}
    \caption{\textbf{Overview of the motivation of our framework, Learning-to-Adapt-Noise (LAN)}. Instead of adapting a denoising network to unseen noise, LAN adapts the input noise itself by directly learning to offset the deviations between the unseen noise and the noise distribution a denoising network is trained on.}
    \vspace{-1.5em}
    \label{fig:overview}
\end{figure}
While early CNN-based models have brought substantial improvements, such impressive results are only limited to images with known noise distribution.
Since early neural networks are trained on images with fixed and known noise distributions, models fail to generalize to images that contain unseen noise.
Considering that images often contain unseen/unknown noise, due to new environments and camera models, in real-world applications, the incapability of handling such noise is a critical shortcoming that needs to be addressed.

To bridge the gap between the performance on controlled environments and real-world environments, various works have focused on constructing datasets consisting of real-world clean-noisy image pairs~\cite{dnd,Nam,polyu,renoir,SIDD,siddplus}.
For instance, SIDD~\cite{SIDD} dataset, one of widely accepted and used datasets, is constructed by capturing real-world noisy images with various smartphone cameras under different environment (e.g., ISO, shutter speed, and aperture settings).
However, not only does the acquisition of such real-world datasets require laborious human effort, but it remains difficult to achieve the generalization to unseen noise that is substantially different from noise distribution within the train data.

In parallel with efforts in collecting real-world noise dataset, there has been several attempts in improving the performance on real-world noise with different approaches, which can be mainly classified into two methodologies: generative modeling and self-supervised learning.
Generative modeling approaches employ generative models to synthesize realistic noisy images~\cite{danet,cycleisp,noiseflow,c2n,canoisegan,nfsrgbmodeling}.
However, generative modeling approaches require clean images, from which noisy images are conditionally generated.

On the other hand, self-supervised learning based blind denoising approaches~\cite{noise2noise,noise2self,noise2void,quan2020self2self, wang2023noise2info,apbsn, lee2022noisetransfer, mansour2023zero} make use of statistical assumptions that allow models to be trained only using noisy images.
Such self-supervised learning formulations enable the model to be finetuned and adapted to a given image with unseen noise.
While there has been a lot of efforts in designing more practical and effective self-supervision tasks (e.g., loss functions, targets, etc.), the focus has been less on how models are finetuned with given self-supervision tasks.
Few works have attempted to adapt the model to a given noisy image~\cite{lee2020self, gunawan2022test}.

In this work, instead of adapting the model to handle fine-grained deviations in input image that unseen noise brings, we propose to learn to directly offset the deviations between the unseen noise distribution in the input and the noise distribution expected by a pretrained denoising network, as shown in Figure~\ref{fig:overview}.
To this end, we propose a new framework, dubbed Learning to Adapt Noise (LAN), which adapts an input noise with a learnable pixel-wise offset that is trained with the aid of self-supervision tasks.
Orthogonal and complementary to self-supervised learning algorithms, our framework is shown to bring substantial improvements, in comparison to a model adaptation approach. 

\begin{figure*}[t]
    \centering
    \begin{subfigure}[b]{0.22\textwidth}
        \centering\includegraphics[height=6.7cm]{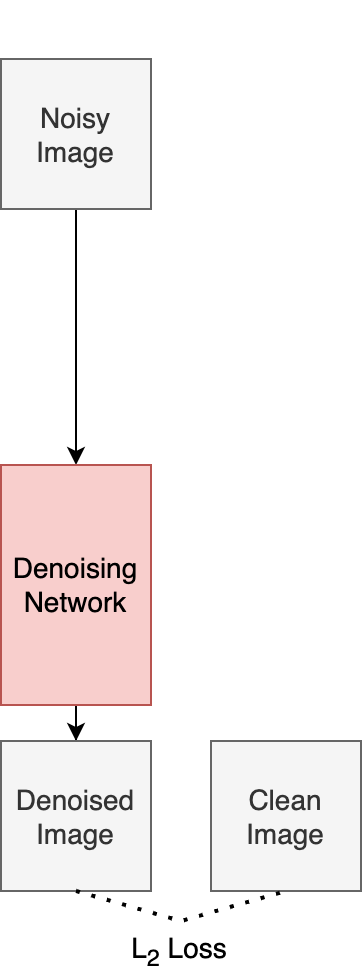}
        \caption{Pretraining}
        \label{method_1}
    \end{subfigure}
    \hspace{0.6cm}
    \begin{subfigure}[b]{0.22\textwidth}
        \centering        
        \includegraphics[height=6.7cm]{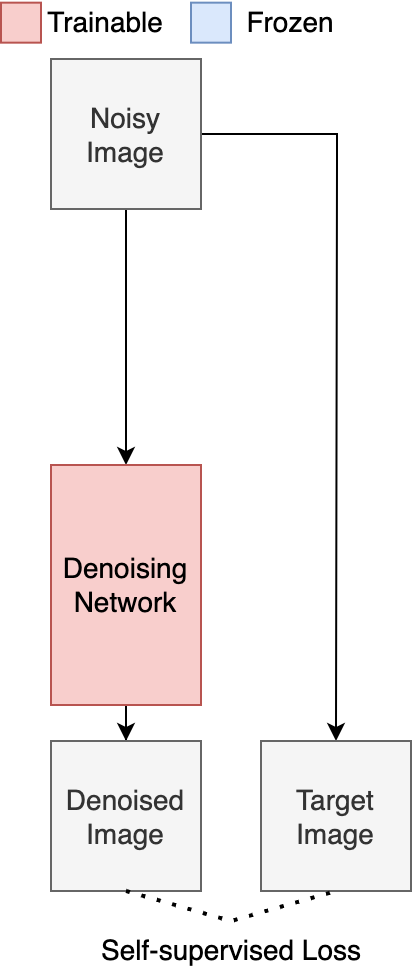}
        \caption{Fine-tuning}
        \label{method_2}
    \end{subfigure}
    \hspace{0.6cm}
    \begin{subfigure}[b]{0.22\textwidth}
        \centering        
        \includegraphics[height=6.7cm]{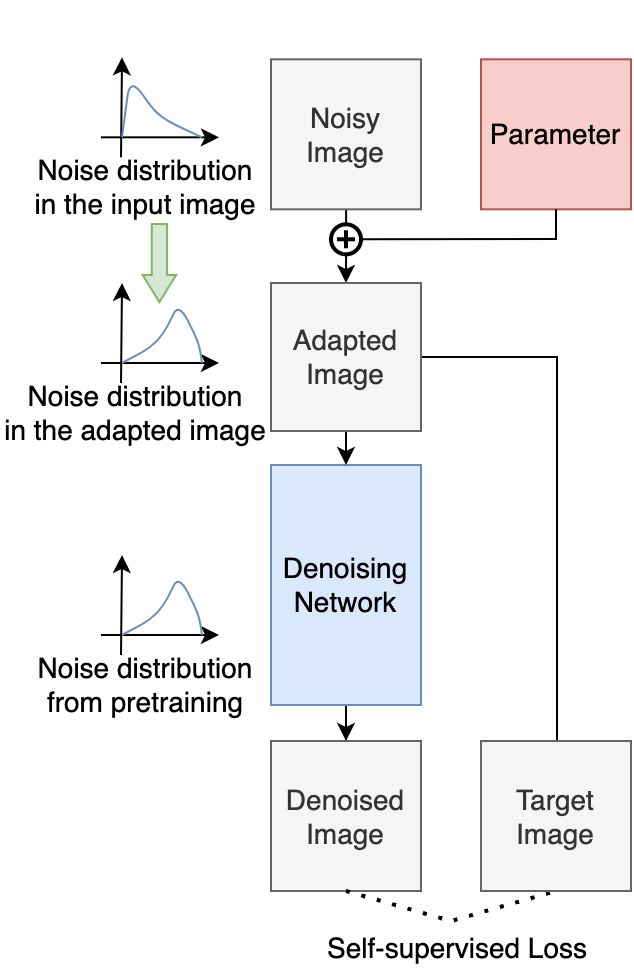}
        \caption{LAN (Ours)}
        \label{method_3}
    \end{subfigure}
 \vspace{-0.5em}
   \caption{
   \textbf{Overview of conventional methods and our framework, Learning-to-Adapt-Noise (LAN, ours).}
   (a) Pretraining of a denoising network is done with pairs of noisy-clean images, with standard L$_2$ loss.
   (b) Fine-tuning of a whole denoising network is done with only a given noisy image via self-supervision loss function, such as ZS-N2N~\cite{mansour2023zero}, to handle unseen noise in the image.
   (c) Learning-to-Adapt-Noise (LAN, ours) is similar to fine-tuning in that only a give noisy image is used with self-supervision loss. 
   However, our method keeps the whole denoising network frozen and only adapts a given noisy image to handle unseen noise.
 }
 \vspace{-1.5em}
   \label{fig:method}
\end{figure*}
\section{Related works}
\label{sec:related_work}
\noindent\textbf{Single image denoising.}
Image denoising is a crucial area of computer vision research.
Early approaches include total variation-based denoising~\cite{rudin1992nonlinear}, sparse coding-based denoising~\cite{mairal2009non}, and self-similarity-based denoising methods~\cite{buades2005non,dabov2007image}. 
With the success of deep learning in computer vision, many deep learning-based denoising methods have been developed, starting with methods that combine sparse coding and MLP~\cite{xie2012image}. 
DnCNN~\cite{dncnn} focuses on noise by residual structure. 
FFDNet~\cite{zhang2018ffdnet} uses downsampling and non-uniform noise level maps to achieve a faster and more efficient performance. 
RIDNet~\cite{anwar2019real} has further improved performance especially on real-world noisy datasets, such as SIDD~\cite{SIDD}, by incorporating a reinforcement attention module that utilizes global information such as feature attention and local skip connection bases.
In recent years, vision transformers~\cite{vit} have brought significant advancements in image restoration. 
Several works have employed vision transformers to bring futher improvements to denoising~\cite{chen2021pre, liang2021swinir, restormer, Uformer_Wang_2022_CVPR}.

\noindent\textbf{Blind image denoising.} While denoising neural networks achieve high performance, they require pairs of clean and noisy images.
However, obtaining clean images can be challenging.
Several blind denoising approaches~\cite{dip, quan2020self2self, noise2self, wang2023noise2info} have emerged to tackle unseen noise image denoising without the need for clean images to training network. Early strategy in blind denoising is based on an internal image prior~\cite{dip}. Recently, blind denoising approaches based on self-supervised learning have endeavored to achieve performance levels comparable to supervised learning.
Under the assumption that noise in each image is independent and has a zero-mean distribution, Noise2Noise~\cite{noise2noise} takes different images of a scene that act as input and target images for training a network, respectively. 
Noise2Self~\cite{noise2self} is another self-supervised learning method that masks the noisy image at regular intervals, using the remaining pixels as an input image and masked pixels as a target image. On the other hand, recent works have focused on creating a input-target pair by subsampling from a single noisy image. 
Notably, Neighbor2Neighbor~\cite{huang2021neighbor2neighbor} is uses a random neighbor sub-sampler to generate input-target pairs for training a network.
Zero-Shot Noise2Noise~\cite{mansour2023zero} is another self-supervised learning approach that generates input-target pairs by applying a filter that computes the mean of diagonal pixels. 
By test-time adaptation (TTA)~\cite{kundu2020universal, yang2021generalized, wang2021tent} using these self-supervised learning approaches, a pretrained denoising network can be either finetuned to handle unseen noise~\cite{lee2020self, gunawan2022test}.
However, the misalignment between new unseen noise and noise expected by a pretrained denoising network may lead to suboptimal performance even after adaptation.
In this work, we approach the problem from a different perspective: adapt a new noisy image itself to reduce the misalignment itself.
Figure~\ref{fig:method} outlines the major differences between our framework LAN and standard approaches.

\noindent\textbf{Domain adaptive image translation.}
Adaptation of an input noisy image to handle the misalignment shares motivations with image translation for domain adaptation.
Image translation aims to learn to translate images from a source domain to a target domain, through the aid of generative models~\cite{imagetranslation:kingma2013auto,goodfellow2014generative,zhu2017toward, liu2017unsupervised, lee2018diverse, isola2017image, ma2018exemplar}.
While our proposed framework shares some similarities with these works from the perspective of domain adaptation, domain adaptation approaches either require a large amount of target-domain images or require a large amount of images from various domains in order to train a large generative model for image translation.
On the other hand, we do not need a separate generative model for image translation.
We also do not assume the availability of a large amount of images with new unseen noise distribution.
In fact, we train a learnable offset that is directly added to each input image for adaptation, without the need for an additional network for image translation or the need for other images.
Furthermore, we believe it is an interesting perspective to draw connections between image denoising and domain adaptation.

\noindent\textbf{Adversarial attack on image domain.}
\label{rw:adv}
Directly optimizing the offset added to an image exhibit similar characteristics as adversarial attack.
Adversarial attack is often performed via adversarial examples generated by adding a learnable noise to the input, where noise is optimized to make the neural network output deviate from its original value~\cite{adversarial:carlini2017towards,adversarial:szegedy2013intriguing}.
However, while adversarial attacks aim to degrade the model performance, our work focuses on improving the performance.
Regardless, we introduce a new interesting perspective, where denoising performance can be improved with learnable anti-adversarial noise.
\vspace{-0.5em}
\section{Proposed method}
\subsection{Preliminaries}
\noindent\textbf{Problem formulation.}
In this work, we aim to tackle a scenario, where a new input image contains noise that follows a different distribution from the distribution a denoising network is trained with.
Formally, a denoising network $f$ with parameters $\vtheta$ is pretrained with pairs of clean images $\vx^s$ and their noisy counterparts $\vy^s$ that contain noise $\ve^s$ assumed to follow a certain distribution $\gD^s$, constructed as follows:
\begin{equation}
\label{eq:train_noise}
    \vy^s = \vx^s + \ve^s, \qquad \text{where} \quad \ve^s \sim \gD^s.
\end{equation}
In particular, a denoising network $f_\vtheta$ is trained to map noisy images $\vy^s$ with certain noise $\ve^s$ to their clean counterparts $\vx^s$ via minimizing the empirical loss as follows:
\begin{equation}
\label{eq:train_loss}
    \bm{\theta}^* = 
    \argmin_{\bm{\theta}}\mathop{\mathbb{E}}\left[\big\| f_{\bm{\theta}} (\bm{y}^s) - \bm{x}^s \big\|^2_2\right].
\end{equation}

During test time, we expect that a new input noisy image $\vy^u$ contains an unknown clean image $\vx^u$ with unseen noise $\ve^u$ that follows a distribution $\gD^u$ that is different from the distribution seen during training (i.e., $\gD^s$):
\begin{equation}
\label{eq:test_noise}
    \vy^u = \vx^u + \ve^u, \qquad \text{where} \quad \ve^u \sim \gD^u.
\end{equation}
Under such formulation, challenges arise due to the domain misalignment between a noise distribution a denoising network is trained on and a new noise distribution encountered during the test phase.

\noindent\textbf{Self-supervised learning.}
One approach to handle such misalignment would be to adapt a pretrained network to given noisy image.
However, only a noisy image $\vy^u$ is available while a clean image $\vx^u$ is unavailable, making it difficult to train or adapt a denoising network $f_{\vtheta}$ via Equation~\ref{eq:train_loss}.

To extract an underlying clean image from a noisy image, a few recent works have introduced self-supervised learning approaches~\cite{noise2noise,noise2self,mansour2023zero,neighbor2neighbor}, given the following assumption: clean image pixels and noise pixels exhibit different attributes.
Namely, clean image pixels are highly correlated within local regions, whereas the noise pixels are independent.
Upon the assumption, two independent noisy images, $\vy^u_1$ and $\vy^u_2$, are created out of the same scene from the original noisy image $\vy^u$ through two different transformations $D_1$ and $D_2$, such as downsampling~\cite{neighbor2neighbor,mansour2023zero}.
Then, to approximate the optimization in Equation~\ref{eq:train_loss} without clean images, one of two independent noisy images is used as an input to a network while the other as a target:
\begin{equation}
\label{eq:self_loss}
\gL_{\text{self}} =\big\| f_\vtheta(D_1(\vy^u)) - D_2(\vy^u) \big\|^2_2 \,.
\end{equation}

\subsection{Learning to adapt noise (LAN)}
In this work, we aim to utilize a pretrained denoising network $f_{\vtheta^*}$ to remove new unseen noise $\ve^u$ from a given noisy image $\vy^u$.
However, when a denoising network $f_{\vtheta^*}$ is trained with noise $\ve^s \sim \gD^s$ via Equation~\ref{eq:train_loss} while a new noisy image contains noise $\ve^u \sim \gD^u$, there arises misalignment between new input and what a pretrained network expects.
Such misalignment may worsen as the difference between seen noise distribution $\gD^s$ and unseen noise distribution $\gD^u$ is larger.
The misalignment may lead to suboptimal performance, when directly finetuning a pretrained denoising network $f_{\vtheta^*}$ to minimize self-supervised loss function (Equation~\ref{eq:self_loss}) with new noisy images.

To resolve the noise misalignment issues, we shift the attention to the noise itself at the input level.
In particular, we formulate a new unseen noise as deviation from the seen noise distribution:
\begin{equation}
    \ve^u = \ve^s + \bm{\epsilon}^{s \rightarrow u},
\end{equation}
where $\bm{\epsilon}^{s \rightarrow u}$ represents how much $\ve^u$ deviates from an arbitrary noise $\ve^s$ sampled from $\gD^s$.
Thus, a new noisy image $\vy^u$ can be seen as follows:
\begin{align}
    \vy^u &= \vx^u + \ve^u \\
          &= \vx^u + \ve^s + \bm{\epsilon}^{s \rightarrow u}.
\end{align}
Given the formulation, we observe that we can mitigate the misalignment issues if we can adapt a given noisy image $\vy^u$ to its translated counterpart noisy image $\vy^{u \rightarrow s}$ with seen noise $\ve^s\sim\gD^s$, by removing the deviations $\bm{\epsilon}^{s \rightarrow u}$ as:
\begin{align}
    \vy^{u \rightarrow s} & \coloneqq \vx^u + \ve^s \\
    &= \vx^u + \ve^s + \bm{\epsilon}^{s \rightarrow u} - \bm{\epsilon}^{s \rightarrow u} \\
    &= \vy^u - \bm{\epsilon}^{s \rightarrow u}.
\end{align}

Thus, our goal becomes to find a deviation offset $ - \bm{\epsilon}^{s \rightarrow u}$ that we can add to a given noisy image $\vy^u$ to adapt underlying noise towards noise a pretrained network is more familiar with.
To this end, we add a learnable parameter $\bm{\phi}$ to a given noisy image $\vy^u$ and then train $\bm{\phi}$ to approximate the deviation offset $ - \bm{\epsilon}^{s \rightarrow u}$, as also illustrated in Figure~\ref{fig:overview}:
\begin{equation} 
\label{eq:noise_adapt}
    \vy^{u \rightarrow s} \approx   \vy^u + \bm{\phi}\,.
\end{equation}
However, we do not have access to such deviation offset during the test phase, as new noise distribution is generally unknown and it would be also difficult to explicitly model seen noise distribution.
To approximate an unknown $ - \bm{\epsilon}^{s \rightarrow u}$ with $\bm{\phi}$, we train $\bm{\phi}$ to minimize a self-supervision loss function (Equation~\ref{eq:self_loss}), under the assumption that loss function is minimized when an input noise becomes closer to seen noise distribution.
The assumption is reasonable as a self-supervision loss function is a surrogate of Equation~\ref{eq:train_loss}, which a pretrained network $f_{\vtheta^*}$ is trained to minimize with seen noise distribution.
Overall, our objective function to train $\bm{\phi}$ becomes:
\vspace{-0.1cm}
\begin{equation}
    \label{eq:LAN}
    \bm{\phi}^* = \argmin_{\bm{\phi}}\big\| f_{\vtheta^*}(D_1(\vy^u + \bm{\phi})) - D_2(\vy^u + \bm{\phi}) \big\|^2_2 \,.
\vspace{-0.1cm}
\end{equation}
Note that we freeze the parameters of a pretrained denoising network $\vtheta^*$ and only optimize $\bm{\phi}$.
Then, finally, a clean image is estimated by
\begin{equation}
    \hat{\vx}^u = f_{\vtheta^*}(\vy^u + \bm{\phi}^*).
\vspace{-0.1cm}
\end{equation}

\section{Experiments}
We perform experiments on scenarios of test noisy images with different noise from the training set, demonstrating the effectiveness of our LAN framework. The implementation details and experimental settings are in Section~\ref{sec:exp_details}. Then, we present the results in Section~\ref{sec:results}. The discussions on zero-shot denoising, computational efficiency, and the effects of noise adaptation are covered in Sections ~\ref{sec:zeroshot}, ~\ref{sec:efficiency}, and ~\ref{sec:effect}, respectively.

\subsection{Experimental setup}
\label{sec:exp_details}
\noindent\textbf{Dataset.} For training denoising networks, we use one of widely used real-world noise dataset SIDD~\cite{SIDD}.
To simulate unseen new noise scenarios during the test phase, we utilize other real-world noise datasets with noise distribution different from SIDD: in this work, PolyU~\cite{polyu} and Nam~\cite{Nam}. We split the original PolyU images into $512\times512$ patches and crop them into $256\times256$. For Nam, we use patches splited into $256\times256$.
\\
\noindent\textbf{Models.} We conduct experiments on prominent denoising networks, DnCNN~\cite{dncnn}, Restormer~\cite{restormer}, and Uformer~\cite{Uformer_Wang_2022_CVPR}.
For Restormer and Uformer, we employ the parameters of SIDD-pretrained networks that are publicly available.
On other hand, DnCNN does not have publicly available SIDD-pretrained network parameters.
Thus, we train DnCNN from scratch using the SIDD dataset.
\\
\noindent\textbf{Evaluation.} We measure PSNR and SSIM to evaluate the performance of SIDD-pretrained networks and our LAN framework on noisy images from PolyU and Nam datasets, against four alternative adaptation methods: adapting a whole network (`full-trainable'), adapting only the first layer (`first-layer'), adapting only the last layer (`last-layer'),  and adapting a whole network via meta-learning using first-order MAML~\cite{maml, firstorderMAML} (`meta-learning'). For `meta-learning', we initialize the network from SIDD pretrained weights. Then we train with 200 samples and validate with 100 samples from the SIDD training and validation set.
Such test-time adaptation is conducted individually for each image, using only noisy images.
The adaptation runs for up to 20 iterations and is optimized through Adam optimizer~\cite{adam} with $\beta_1 = 0.9 $ and $\beta_2 = 0.999$. 
At this time, the learning rate of LAN is 5e-4.
For alternative adaptation schemes, we use 5e-6 for `full-train', 5e-4 for `first-layer' and 1e-4 for `last-layer'. The `meta-learning' is with 1e-5 learning rate on both adaptation and meta update.
The learning rates are empirically found for stable convergence of models trained with self-supervision loss functions.

\begin{table*}
  \centering
\vspace{-2em}
  \aboverulesep=0ex
    \belowrulesep=0ex
    \resizebox{0.9\textwidth}{!}{
  \begin{tabular}{cc cc cc cc }
    \toprule
     \multirow{3}{*}{Model}  & \multirow{3}{*}{Method} & \multirow{3}{*}{Iter.} & \multicolumn{2}{c}{SIDD \text{$\rightarrow$} PolyU }& \phantom{abc}&\multicolumn{2}{c}{SIDD \text{$\rightarrow$} Nam } \\
      \cmidrule{4-5}\cmidrule{7-8}
     &  & & \multicolumn{2}{c}{PSNR$^{\uparrow}$ (dB) / SSIM$^{\uparrow}$}& \phantom{abc}&\multicolumn{2}{c}{PSNR$^{\uparrow}$ (dB) / SSIM$^{\uparrow}$} \\
    \cmidrule{4-5}\cmidrule{7-8} & & & ZS-N2N & Nbr2Nbr &&  ZS-N2N & Nbr2Nbr \\
    \midrule

    \multirow{15}{*}{DnCNN} 
    & pretrained & - & \multicolumn{2}{c}{38.10  / 0.952 } & &\multicolumn{2}{c}{36.60  / 0.930 } \\
    \cmidrule{2-5}\cmidrule{7-8}
    
    & \multirow{3}{*}{full-trainable}
    & 5 & 38.07  / 0.951  & 38.08  / 0.951 &
        & 36.60   / 0.929 & 36.60   / 0.929  \\
    && 10 & 38.04  / 0.950  & 38.06   / 0.951  & 
        & 36.59   / 0.928 & 36.60  / 0.928  \\
    && 20 & 37.99   / 0.949  & 38.02  / 0.949  &
        & 36.56  / 0.925  & 36.56  / 0.925   \\
    \cmidrule{2-5}\cmidrule{7-8}
    
    & \multirow{3}{*}{first-layer} 
    & 5 & 37.93  / 0.948  & 37.95  / 0.948 &
        & 36.48   / 0.923 & 36.46   / 0.923  \\
    && 10 & 37.76  / 0.943  & 37.83   / 0.945  & 
        & 36.29   / 0.915 & 36.28   / 0.915  \\
    && 20 & 37.47   / 0.935  & 37.67  / 0.941  &
        & 35.95  / 0.902  & 36.02  / 0.904   \\
    \cmidrule{2-5}\cmidrule{7-8}
    
    & \multirow{3}{*}{last-layer} 
    & 5 & 38.12  / 0.952  & 38.13  / 0.952 &
        & 36.69  / 0.931 & 36.70  / 0.931  \\
    && 10 & 38.13  / 0.952  & 38.14  / 0.952  & 
        & 36.75  / 0.931 & 36.77  / 0.931  \\
    && 20 & 38.13  / 0.952  & 38.14  / 0.952  &
        & \underline{36.78}  / 0.930  & \textbf{36.81}  / 0.930   \\
    \cmidrule{2-5}\cmidrule{7-8}
    
    & \multirow{3}{*}{meta-learning} 
    & 5 & \underline{38.23}  / \textbf{0.955}  & 38.23  / \textbf{0.956} &
        & 36.56   / 0.936 & 36.54   / \textbf{0.935}  \\
    && 10 & \underline{38.23}  / \textbf{0.955}  & \underline{38.25}   / \underline{0.955}  & 
        & 36.66   / 0.934 & 36.65   / \underline{0.934}  \\
    && 20 & 38.17   / 0.953  & 38.20  / 0.954  &
        & 35.69  / 0.931  & 36.68  / 0.930   \\
    \cmidrule{2-5}\cmidrule{7-8}
    
    & \multirow{3}{*}{LAN (Ours)} 
    & 5 & 38.22  / \underline{0.954}  & 38.16  / 0.953 &
        & 36.73   / 0.934 & 36.66   / 0.932  \\
    && 10 & \textbf{38.29}  / \textbf{0.955} & 38.22   / 0.954  & 
        & \textbf{36.79}  / 0.936 & 36.71   / 0.933  \\
    && 20 & \textbf{38.29} / \textbf{0.955}  & \textbf{38.31}  / \textbf{0.956}  &
        & \underline{36.78} / \textbf{0.938}  & \underline{36.80}  / \textbf{0.935}   \\
    \midrule

    \multirow{15}{*}{Restormer} 
    & pretrained & - & \multicolumn{2}{c}{39.03  / 0.966 } & &\multicolumn{2}{c}{38.03  / 0.951 } \\
    \cmidrule{2-5}\cmidrule{7-8}
    
    & \multirow{3}{*}{full-trainable}
    & 5 & 39.09  / 0.966  & 39.04  / 0.965 &
        &  38.14 / 0.952 & 38.07   / 0.951  \\
    && 10 & 39.12  / 0.965  & 39.04   / 0.965  & 
        & 38.23 / 0.952 & 38.08  / 0.950  \\
    && 20 & 39.14   / 0.965  & 38.98  / 0.964  &
        & 38.35 / 0.953  & 38.05  / 0.948   \\
    \cmidrule{2-5}\cmidrule{7-8}
    
    & \multirow{3}{*}{first-layer} 
    & 5 & 39.04  / 0.965  & 39.00  / 0.965 &
        & 38.12 / 0.951 & 38.07   / 0.950   \\
    && 10 & 38.96  / 0.964  & 38.89   / 0.964  & 
        & 38.05 / 0.950 & 37.95   / 0.948  \\
    && 20 & 38.74   / 0.961  & 38.66  / 0.961  &
        & 37.65 / 0.943 & 37.52  / 0.941   \\
    \cmidrule{2-5}\cmidrule{7-8}
    
    & \multirow{3}{*}{last-layer} 
    & 5 & 39.07  / 0.965  & 39.08  / 0.965 &
        & 38.09  / 0.951 & 38.10  / 0.951  \\
    && 10 & 39.06  / 0.965  & 39.07  / 0.965  & 
        & 38.12  / 0.950 & 38.14  / 0.950  \\
    && 20 & 39.02  / 0.964  & 39.03  / 0.964  &
        & 38.12  / 0.948  & 38.14  / 0.948   \\
    \cmidrule{2-5}\cmidrule{7-8}
    
    & \multirow{3}{*}{meta-learning} 
    & 5 & 39.12  / 0.966  & 39.12  / 0.966 &
        & 38.15 / 0.954 & 38.15   / 0.953   \\
    && 10 & 39.18  / 0.966  & 39.13   / 0.966  & 
        & 38.34 / 0.954 & 38.21   / 0.952  \\
    && 20 & 39.19   / 0.965  & 39.06  / 0.964  &
        & 38.49 / 0.954 & 38.17  / 0.949   \\
    \cmidrule{2-5}\cmidrule{7-8}
    
    & \multirow{3}{*}{LAN (Ours)} 
    & 5 & 39.23  / \underline{0.968}  & 39.09  / \underline{0.967} &
        & 38.31 / 0.957 & 38.14   / 0.953  \\
    && 10 & \textbf{39.30}  / \textbf{0.969}  & \underline{39.14} / \underline{0.967}  & 
        & \underline{38.58} / \underline{0.961} & \underline{38.25}   / \underline{0.955}  \\
    && 20 & \underline{39.28}   / \textbf{0.969}  & \textbf{39.17}  / \textbf{0.968}  &
        & \textbf{38.86} / \textbf{0.965} & \textbf{38.38}  / \textbf{0.958}   \\
    \midrule

    \multirow{15}{*}{Uformer} 
    & pretrained & - & \multicolumn{2}{c}{38.93  / 0.965 } & &\multicolumn{2}{c}{37.55  / 0.950 } \\
    \cmidrule{2-5}\cmidrule{7-8}
    
    & \multirow{3}{*}{full-trainable}
    & 5 & 39.01  / 0.964  & 38.96 / 0.964 &
        & 37.80 / 0.950 & 37.72   / 0.948  \\
    && 10 & 39.01  / 0.963  & 38.92   / 0.963  & 
        & 37.97 / 0.950 & 37.77   / 0.946  \\
    && 20 & 38.91   / 0.961  & 38.77  / 0.961  &
        & 38.07 / 0.948 & 37.67  / 0.942   \\
    \cmidrule{2-5}\cmidrule{7-8}
    
    & \multirow{3}{*}{first-layer} 
    & 5 & 38.89  / 0.965  & 38.85  / 0.965 &
        & 37.75 / 0.952 & 37.69   / 0.950  \\
    && 10 & 38.82  / 0.964  & 38.78   /  0.964  & 
        & 37.76 / 0.951 & 37.65   / 0.948  \\
    && 20 & 38.71   / 0.962  & 38.69  /  0.963  &
        & 37.71 / 0.946  & 37.54  / 0.943   \\
    \cmidrule{2-5}\cmidrule{7-8}
    
    & \multirow{3}{*}{last-layer} 
    & 5 & 38.98  / 0.965  & 38.99  / 0.965 &
        & 37.68  / 0.949 & 37.69  / 0.950  \\
    && 10 & 39.00  / 0.965  & 39.01  / 0.965  & 
        & 37.79  / 0.949 & 37.81  / 0.949  \\
    && 20 & 39.01  / 0.965  & 39.01  / 0.965  &
        & 37.90  / 0.949  & 37.92  / 0.949   \\
    \cmidrule{2-5}\cmidrule{7-8}
    
    & \multirow{3}{*}{meta-learning} 
    & 5 & 39.10  / \underline{0.967}  & 39.09  / \textbf{0.967} &
        & 37.77 / 0.957 & 37.87   / \underline{0.955}  \\
    && 10 & \underline{39.20}  / 0.966  & \textbf{39.11}   /  0.966  & 
        & 38.26 / 0.957 &\textbf{38.07}  / 0.952  \\
    && 20 & 39.11   / 0.964  & 38.97  /  0.963  &
        & \textbf{38.52} / 0.956  & 38.00  / 0.947   \\
    \cmidrule{2-5}\cmidrule{7-8}
    
    & \multirow{3}{*}{LAN (Ours)} 
    & 5 & 39.12  / \underline{0.967}  & 39.00  / \underline{0.966} &
        &  37.82 / 0.955 & 37.69   / 0.951  \\
    && 10 & \textbf{39.21}  / \textbf{0.968}  & 39.05   / \underline{0.966}  & 
        & 38.09 / 0.960 & 37.83   / 0.953  \\
    && 20 & \underline{39.20} / \textbf{0.968}  & \underline{39.10}  / \textbf{0.967}  &
        & \underline{38.36} / \textbf{0.964} & \underline{38.02}  / \textbf{0.956}   \\
    \bottomrule
  \end{tabular}
  }
  \caption{
  \textbf{Quantitative comparison on denoising performance} for each combination of a denoising network backbone, an adaptation method, and a self-supervision loss on the real-world noise datasets (PolyU and Nam) after pretraining on another real-world noise dataset, SIDD.
  }
  \vspace{-1em}
  \label{tab:main}
\end{table*}

\def\imgsize{0.16} 
\begin{figure*}
 \begin{subfigure}[b]{\imgsize\textwidth}
        \centering
        \includegraphics[width=\textwidth]{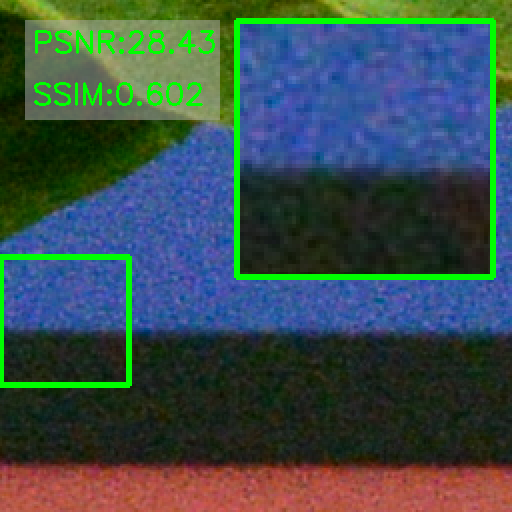}
    \end{subfigure}
    \centering
    \begin{subfigure}[b]{\imgsize\textwidth}
        \centering
        \includegraphics[width=\textwidth]{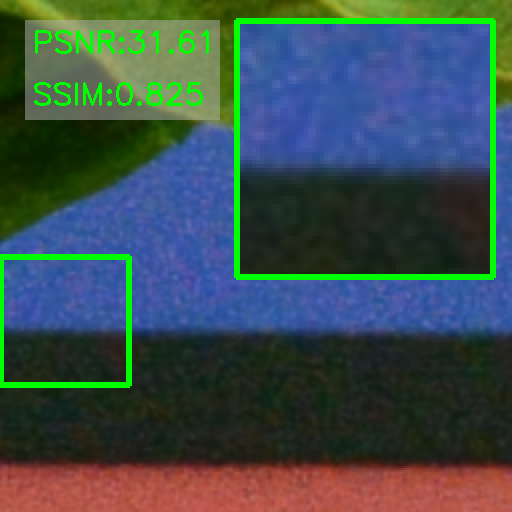}
    \end{subfigure}
    \centering
    \begin{subfigure}[b]{\imgsize\textwidth}
        \centering
        \includegraphics[width=\textwidth]{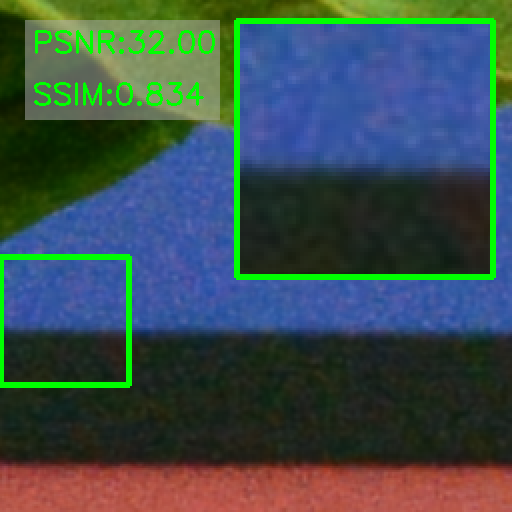}
    \end{subfigure}
    \centering
    \begin{subfigure}[b]{\imgsize\textwidth}
        \centering
        \includegraphics[width=\textwidth]{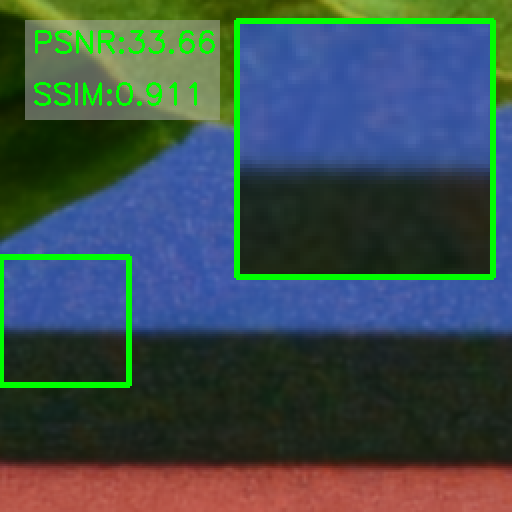}
    \end{subfigure}
    \centering
    \begin{subfigure}[b]{\imgsize\textwidth}
        \centering
        \includegraphics[width=\textwidth]{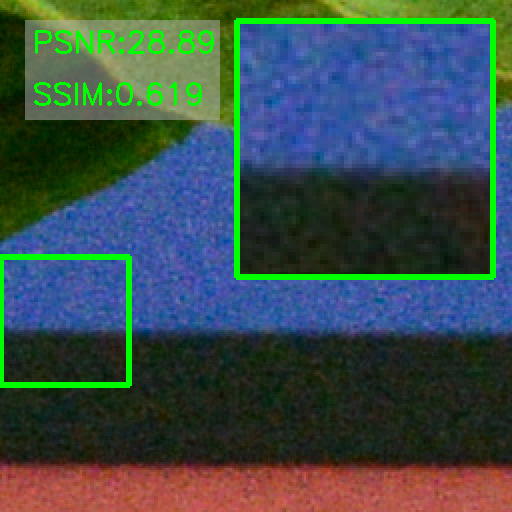}
    \end{subfigure}
    \centering
    \begin{subfigure}[b]{\imgsize\textwidth}
        \centering
        \includegraphics[width=\textwidth]{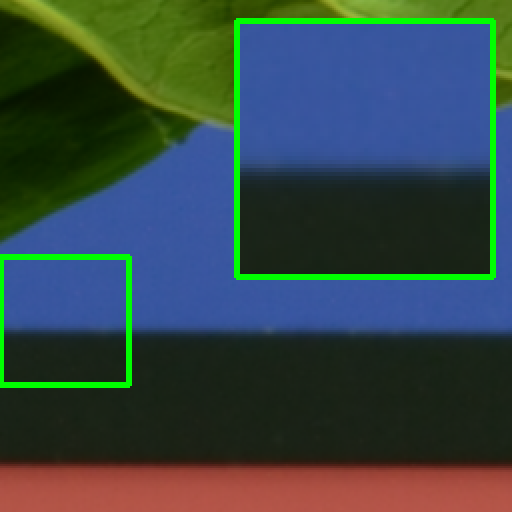}
    \end{subfigure}
    \\
    \vspace{0.5em}
    \begin{subfigure}[b]{\imgsize\textwidth}
        \centering 
        \includegraphics[width=\textwidth]{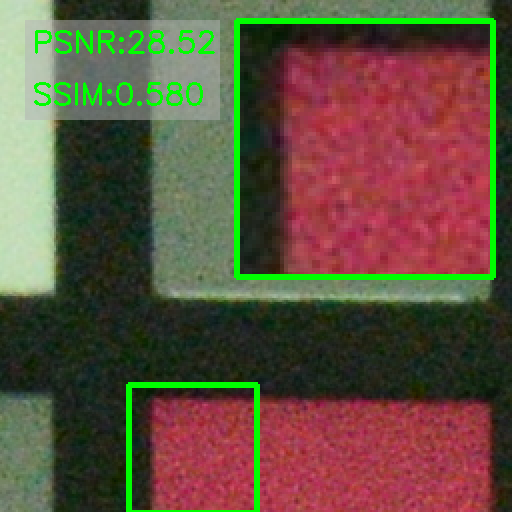}
    \end{subfigure}
    \begin{subfigure}[b]{\imgsize\textwidth}
        \centering
        \includegraphics[width=\textwidth]{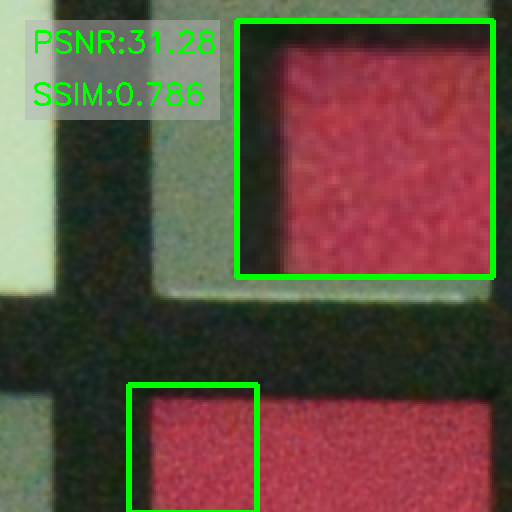}
    \end{subfigure}
    \centering
    \begin{subfigure}[b]{\imgsize\textwidth}
        \centering
        \includegraphics[width=\textwidth]{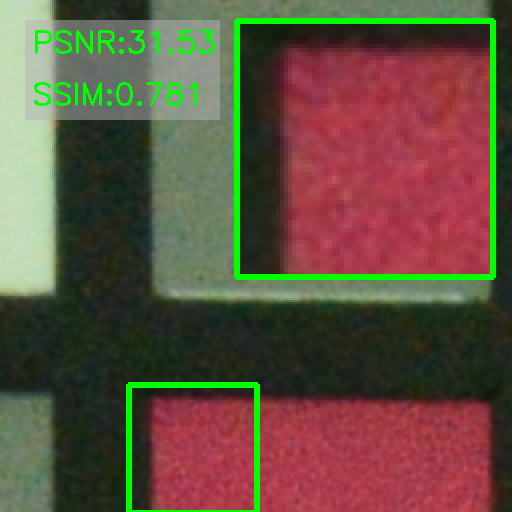}
    \end{subfigure}
    \centering
    \begin{subfigure}[b]{\imgsize\textwidth}
        \centering
        \includegraphics[width=\textwidth]{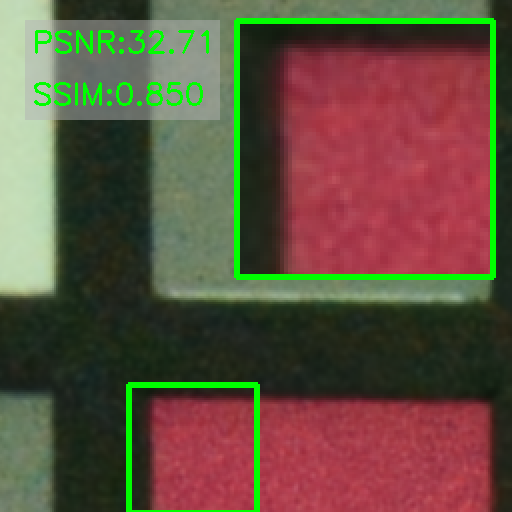}
    \end{subfigure}
    \centering
    \begin{subfigure}[b]{\imgsize\textwidth}
        \centering
        \includegraphics[width=\textwidth]{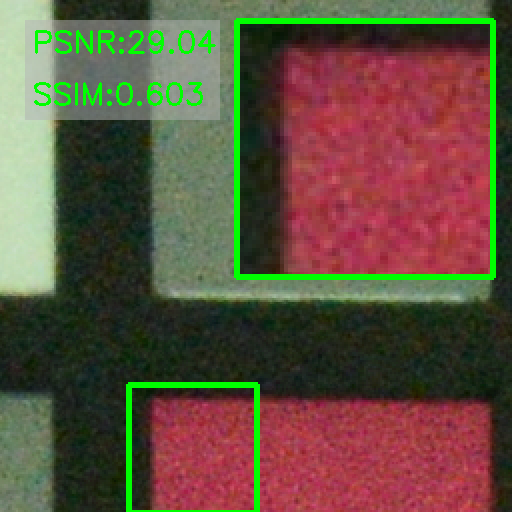}
    \end{subfigure}
    \centering
    \begin{subfigure}[b]{\imgsize\textwidth}
        \centering
        \includegraphics[width=\textwidth]{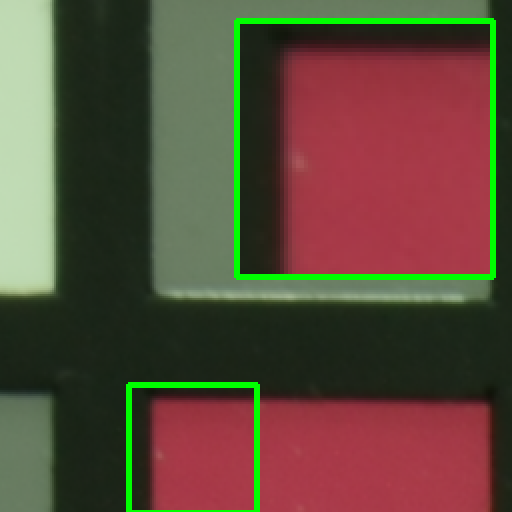}
    \end{subfigure}
    \\
    \vspace{0.5em}
    \begin{subfigure}[b]{\imgsize\textwidth}
        \centering
        \includegraphics[width=\textwidth]{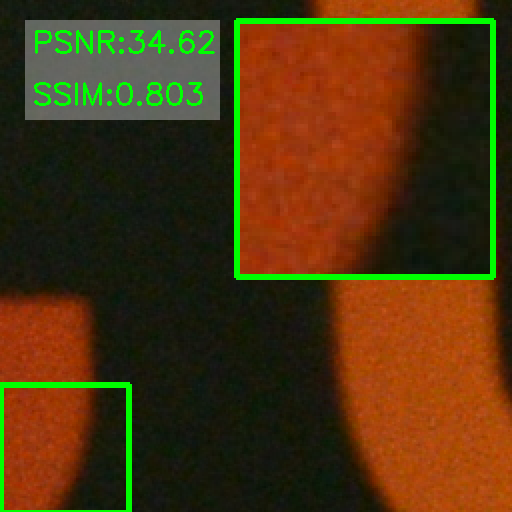}
    \end{subfigure}
    \begin{subfigure}[b]{\imgsize\textwidth}
        \centering
        \includegraphics[width=\textwidth]{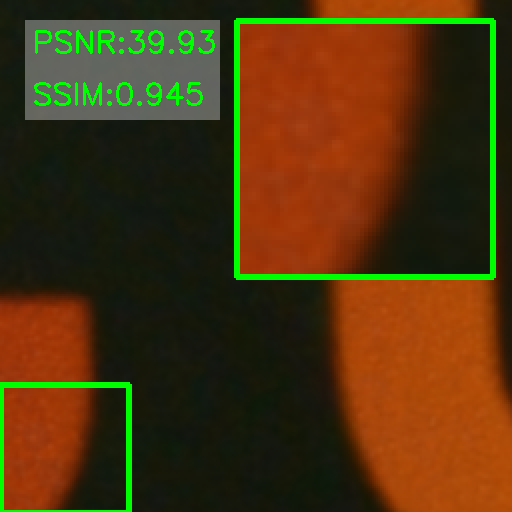}
    \end{subfigure}
    \centering
    \begin{subfigure}[b]{\imgsize\textwidth}
        \centering
        \includegraphics[width=\textwidth]{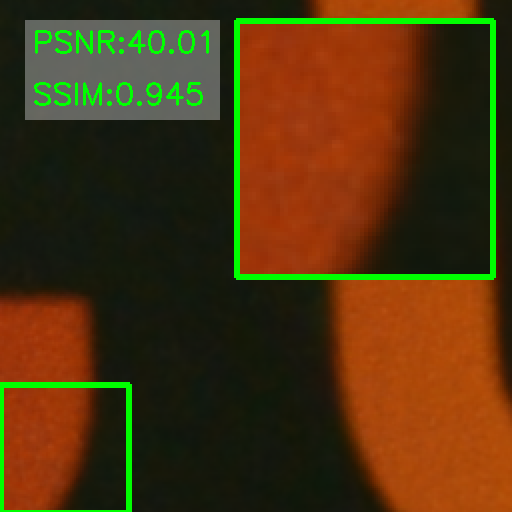}
    \end{subfigure}
    \centering
    \begin{subfigure}[b]{\imgsize\textwidth}
        \centering
        \includegraphics[width=\textwidth]{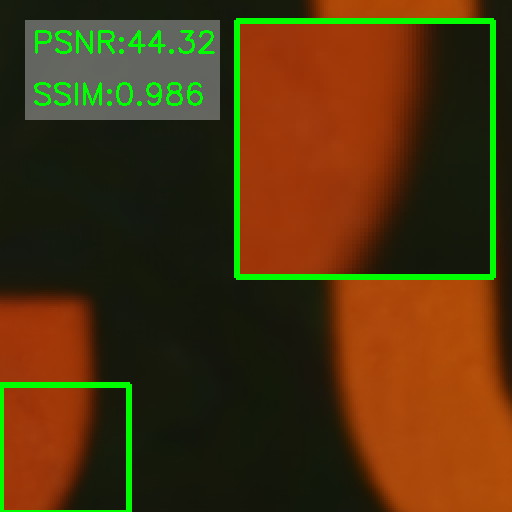}
    \end{subfigure}
    \centering
    \begin{subfigure}[b]{\imgsize\textwidth}
        \centering
        \includegraphics[width=\textwidth]{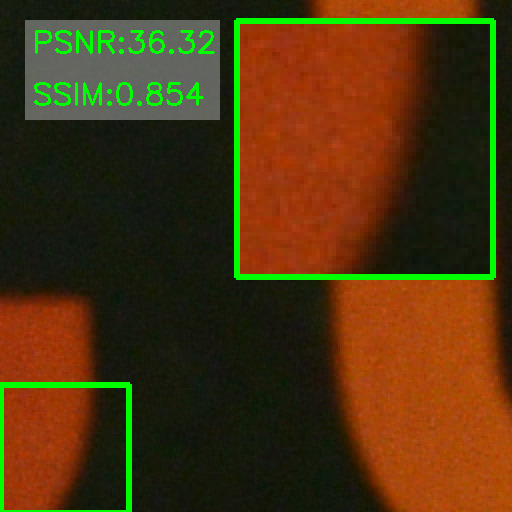}
    \end{subfigure}
    \centering
    \begin{subfigure}[b]{\imgsize\textwidth}
        \centering
        \includegraphics[width=\textwidth]{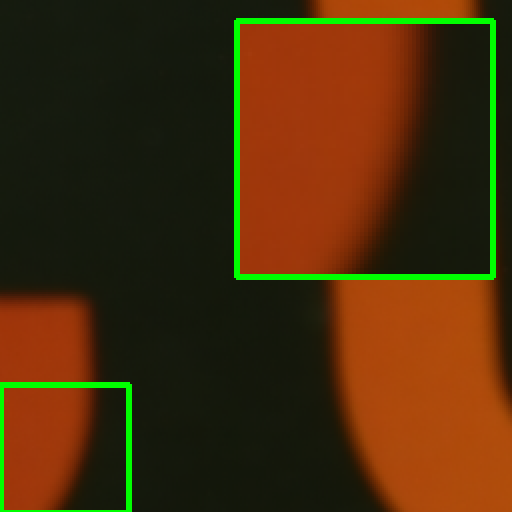}
    \end{subfigure}
    \\
    \vspace{0.5em}
    \begin{subfigure}[b]{\imgsize\textwidth}
        \centering
        \includegraphics[width=\textwidth]{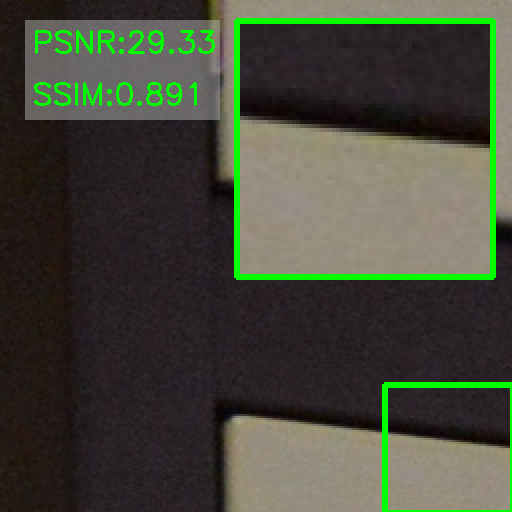}
    \end{subfigure}
    \begin{subfigure}[b]{\imgsize\textwidth}
        \centering
        \includegraphics[width=\textwidth]{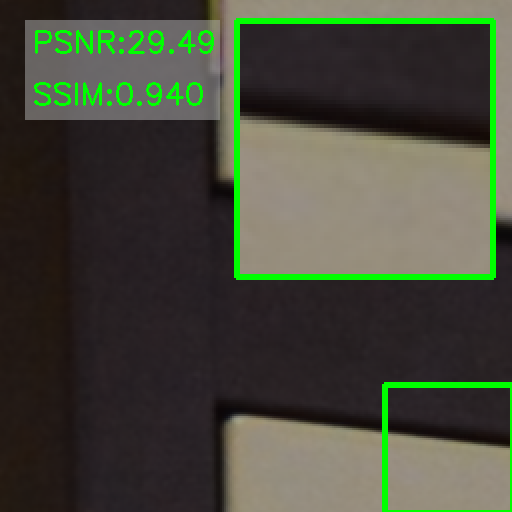}
    \end{subfigure}
    \centering
    \begin{subfigure}[b]{\imgsize\textwidth}
        \centering
        \includegraphics[width=\textwidth]{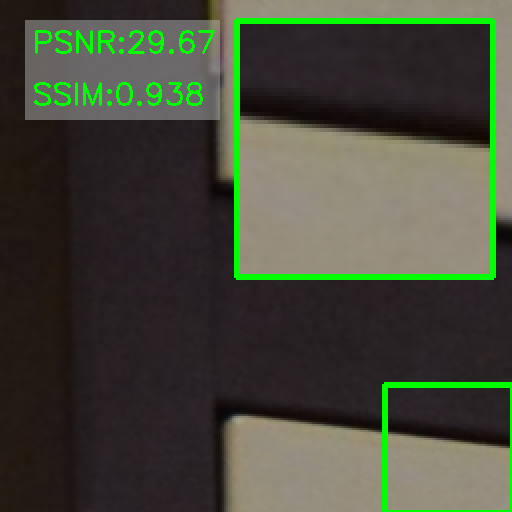}
    \end{subfigure}
    \centering
    \begin{subfigure}[b]{\imgsize\textwidth}
        \centering
        \includegraphics[width=\textwidth]{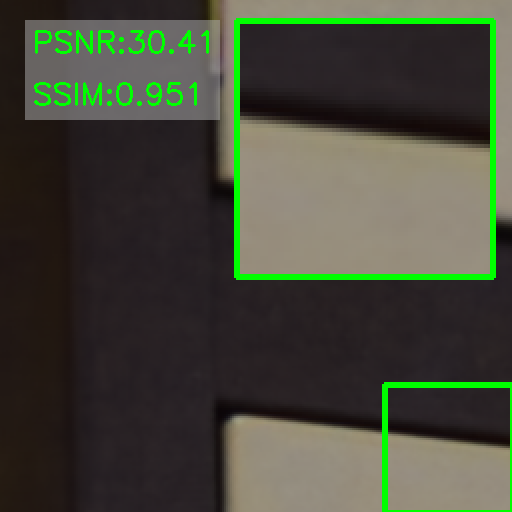}
    \end{subfigure}
    \centering
    \begin{subfigure}[b]{\imgsize\textwidth}
        \centering
        \includegraphics[width=\textwidth]{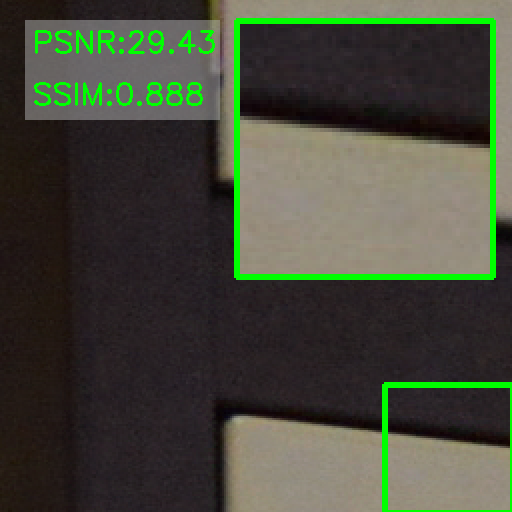}
    \end{subfigure}
    \centering
    \begin{subfigure}[b]{\imgsize\textwidth}
        \centering
        \includegraphics[width=\textwidth]{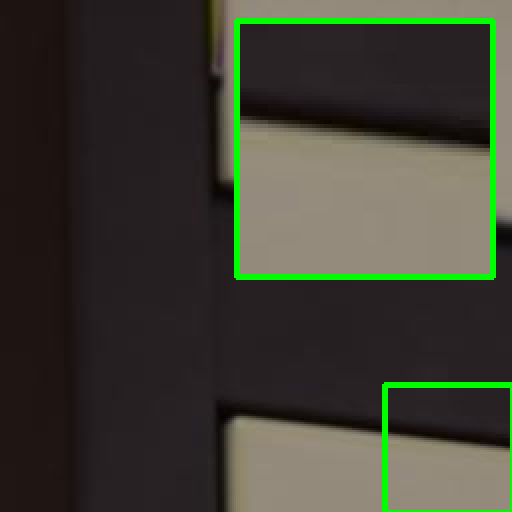}
    \end{subfigure}
    \\
    \vspace{0.5em}
    \begin{subfigure}[b]{\imgsize\textwidth}
        \centering
        \includegraphics[width=\textwidth]{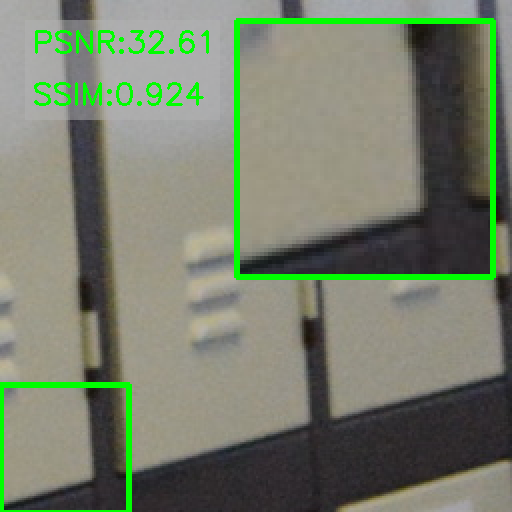}
        \caption*{Noisy image}
        \vspace{1em}
    \end{subfigure}
    \begin{subfigure}[b]{\imgsize\textwidth}
        \centering
        \includegraphics[width=\textwidth]{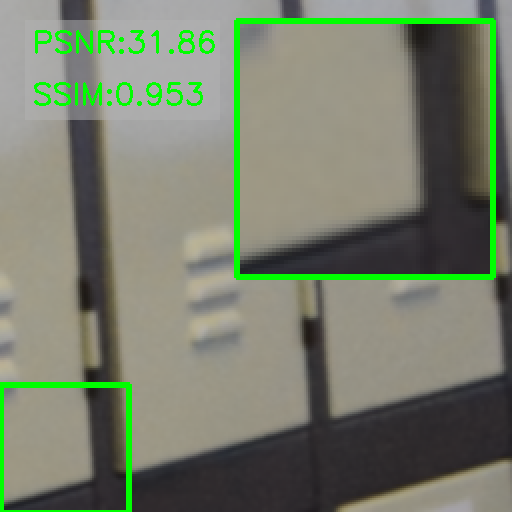}
        \caption*{Pretrain}
        \vspace{1em}
    \end{subfigure}
    \centering
    \begin{subfigure}[b]{\imgsize\textwidth}
        \centering
        \includegraphics[width=\textwidth]{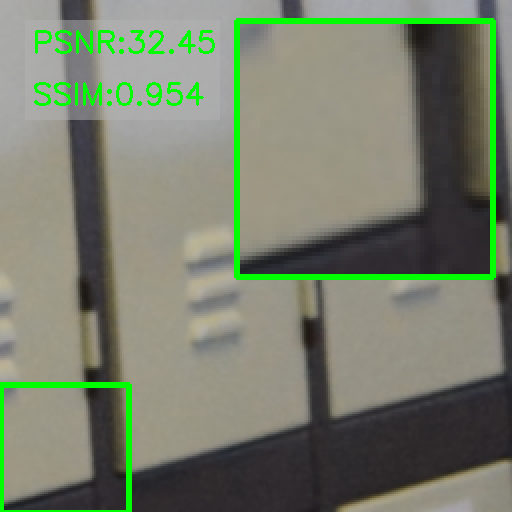}
        \caption*{Full-trainable}
        \vspace{1em}
    \end{subfigure}
    \centering
    \begin{subfigure}[b]{\imgsize\textwidth}
        \centering
        \includegraphics[width=\textwidth]{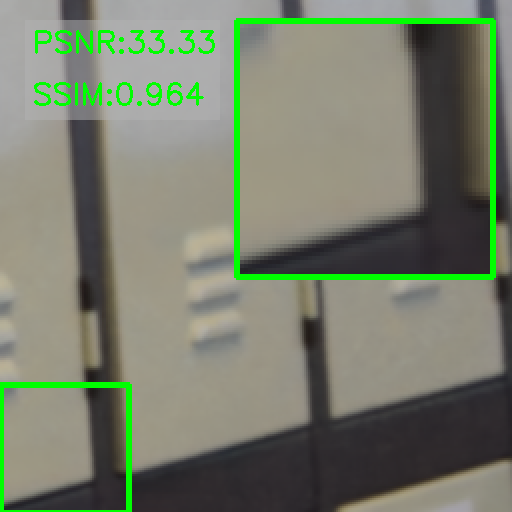}
        \caption*{LAN (Ours)}
        \vspace{1em}
    \end{subfigure}
    \centering
    \begin{subfigure}[b]{\imgsize\textwidth}
        \centering
        \includegraphics[width=\textwidth]{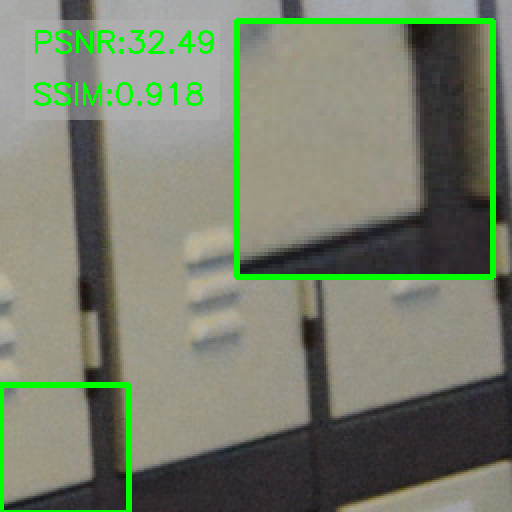}
        \caption*{\centering\shortstack{Adapted noisy image \\ by LAN}}
    \end{subfigure}
    \centering
    \begin{subfigure}[b]{\imgsize\textwidth}
        \centering
        \includegraphics[width=\textwidth]{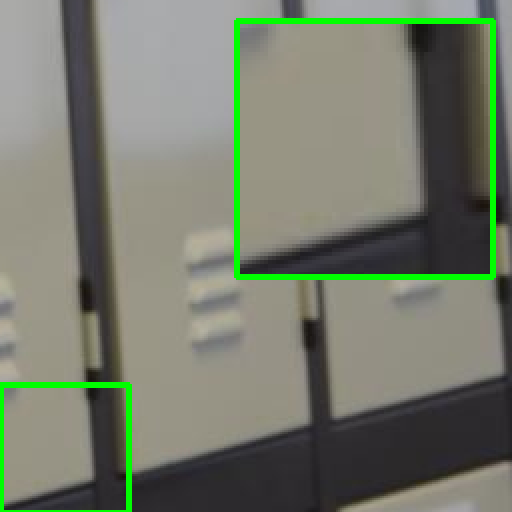}
        \caption*{Clean image}
        \vspace{1em}
    \end{subfigure}
    \\
    \centering
    \vspace{-1em}
    \caption{
    Qualitative comparisons among different adaptation methods.
    Images are obtained with SIDD-pretrained Uformer.
    Full-trainable and LAN (Ours) finetuned the pretrained network via ZS-N2N for 20 iterations on Nam (first three rows) and PolyU (last two rows).
    }
    \vspace{-0.5em}
    \label{fig:qual}
\end{figure*}

\begin{figure*}
    \begin{subfigure}[b]{\imgsize\textwidth}
        \centering
        \includegraphics[width=\textwidth]{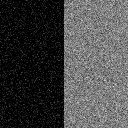}
        \caption*{\scriptsize Noisy image w/ train noise}
    \end{subfigure}
    \centering
    \begin{subfigure}[b]{\imgsize\textwidth}
        \centering
        \includegraphics[width=\textwidth]{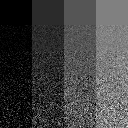}
        \caption*{\scriptsize Noisy image w/ new noise}
    \end{subfigure}
    \centering
    \begin{subfigure}[b]{\imgsize\textwidth}
        \centering
        \includegraphics[width=\textwidth]{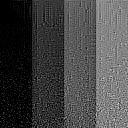}
        \caption*{\fontsize{6}{7}\selectfont Adapted noisy image by LAN}
    \end{subfigure}
    \centering
    \begin{subfigure}[b]{\imgsize\textwidth}
        \centering
        \includegraphics[width=\textwidth]{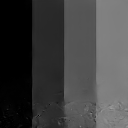}
        \caption*{\scriptsize Pretrained}
    \end{subfigure}
    \centering
    \begin{subfigure}[b]{\imgsize\textwidth}
        \centering
        \includegraphics[width=\textwidth]{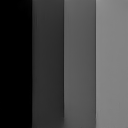}
        \caption*{\scriptsize LAN}
    \end{subfigure}
    \centering
    \begin{subfigure}[b]{\imgsize\textwidth}
        \centering
        \includegraphics[width=\textwidth]{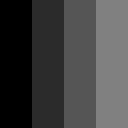}
        \caption*{\scriptsize Clean image}
    \end{subfigure}
    \centering
    \vspace{-0.5em}
    \caption{
    Visualization of synthetic noisy images.
    Noisy image with train noise is a noisy image that is used for pretraining a denoising network (DnCNN).
    Noisy image with new noise contain a new noise that is different from pretraining.
    Adapted noisy image by LAN is a result of noise adaptation of noisy image with new noise.
    We observe that noise in the adapted noisy image becomes more similar to noisy image with train noise.
    Particularly, we observe that noise has been added to the top of the image, where there was previously no noise.
    As a result, LAN helps achieve better denoising performance.
    }
    \vspace{-1em}
    \label{fig:synthetic}
\end{figure*}

\subsection{Experimental results}
\label{sec:results}
\noindent\textbf{Quantitative results.}
We test the adaptation methods in two scenarios: SIDD $\rightarrow$ PolyU and SIDD $\rightarrow$ Nam.
SIDD $\rightarrow$ PolyU means a network is pretrained on SIDD dataset and evaluated on PolyU dataset, while SIDD $\rightarrow$ Nam represents a scenario, where a network is pretrained on SIDD and evaluated on Nam dataset.
To demonstrate the applicability of our framework, we also perform experiments with various network backbones: namely, DnCNN~\cite{dncnn}, Restormer~\cite{restormer}, and Uformer~\cite{Uformer_Wang_2022_CVPR} over different self-supervised losses such as ZS-N2N~\cite{mansour2023zero} and Nbr2Nbr~\cite{huang2021neighbor2neighbor}.
The experimental results are displayed in Table~\ref{tab:main}.

The results demonstrate that LAN exhibits notable performance improvement even after performing adaptation for only 5 iterations.
Furthermore, our framework LAN is shown to consistently outperform a `full-trainable' method that adapts all parameters of a network to a given noisy image with the same self-supervision loss function (Equation~\ref{eq:self_loss}) across problem settings and network backbones.
For instance, under the setting of Restormer with Nbr2Nbr self-supervised loss on SIDD $\rightarrow$ NAM, our LAN framework improves the pretraining performance by $0.35$dB, while  a `full-trainable' method improves by only $0.05$dB.
As a matter of fact, a `full-trainable' method often results in performance degradation, let alone improve the performance of a pretrained network.

One may argue that fine-tuning all parameters of a network may lead to overfitting, hence the reason for the relatively low performance gain.
It can also be argued that our method of directly adjusting the noise in the input image can prevent overfitting and have similar effects to finetuning only the first layer of a network.
However, `full-trainable' adaptation often leads to performance improvements.
Furthermore, finetuning only a first layer (denoted as `first-layer' in the table) results in inferior performance compared to `full-trainable' and our method.
The results suggest that our method does not have similar effects to finetuning only the first layer. In contrast to fine-tuning only the first layer, our method offers fine-grained noise adaptation on a per-pixel basis, rather than relying solely on convolution, which carries a significant inductive bias.

\begin{figure}[t]
    \vspace{-1em}
    \centering
    \includegraphics[width=0.33\textwidth]{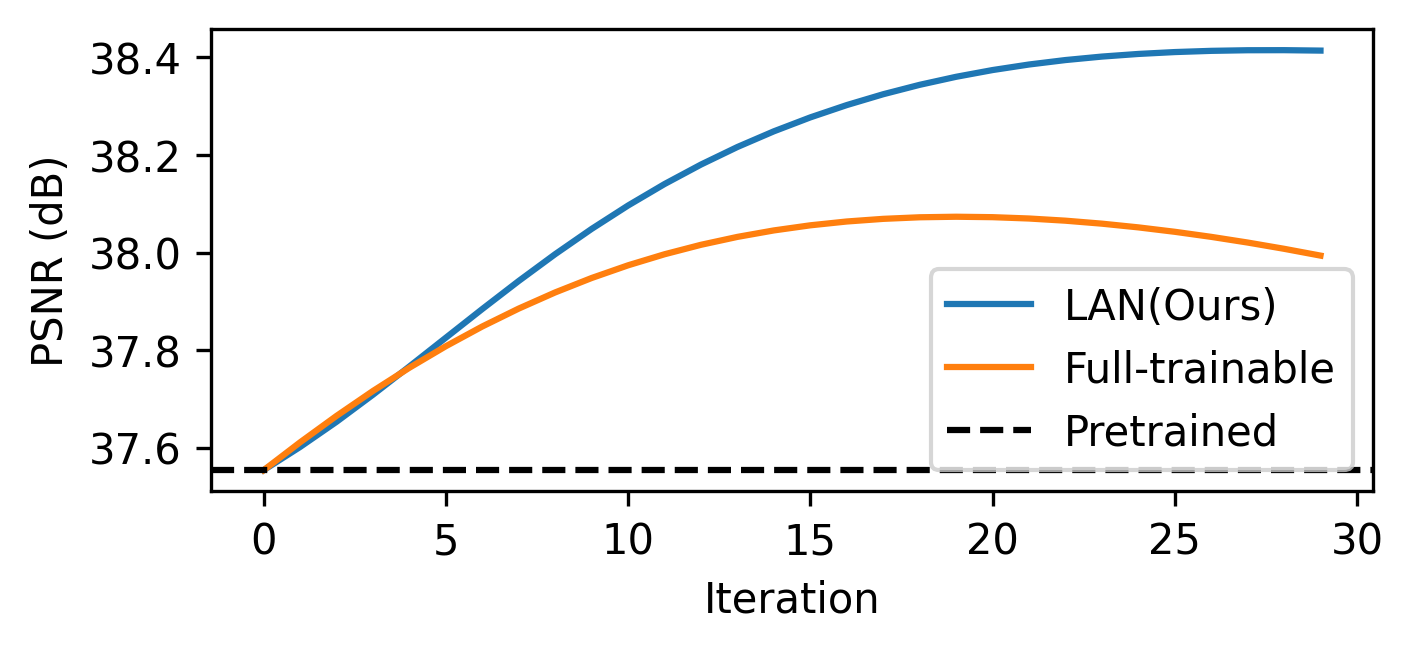}
    \vspace{-1.0em}
    \caption{Plot of performance in PSNR over the number of adaptation iterations. Results are obtained with Uformer finetuned via ZS-N2N on Nam Dataset.}
    \vspace{-0.5em}
    \label{fig:perf_curve}
\end{figure}

On the other hand, the `full-trainable' adaptation can sometimes bring improvements even after 20 iterations, as it can be seen with Uformer network backbone adapted with ZS-N2N loss function on Nam dataset.
Then, one may be curious as to whether `full-trainable' adaptation can outperform LAN if adaptation is performed for longer iterations.
As such, we plot the performance curve of `full-trainable' and our LAN method as the number of iterations increases with \textcolor{black}{Uformer network backbone adapted with ZS-N2N loss function on Nam dataset, as visualized in Figure~\ref{fig:perf_curve}.}
As shown in the figure, even after performing adaptation for longer iterations, `full-trainable' method fails to achieve performance on par with our method and starts to worsen the performance after near 20 iterations. 

\noindent\textbf{Qualitative results.}
We display the qualitative results of a pretrained network, `full-trainable' adaptation, and our LAN framework in Figure~\ref{fig:qual}.
The images are obtained with Uformer finetuned via ZS-N2N for 20 iterations on Nam dataset. 
We also visualize the adapted noisy image by our method.
Interestingly, noise adaptation sometimes has better PSNR/SSIM than an original noisy image.
Then, one may think that the performance of LAN may be because noise adaptation is introducing additional denoising process.
However, the denoising performance improvement by noise adaptation is small, compared to pretrained network and our whole LAN framework.
Furthermore, we observe that noise adaptation itself does not always give better PSNR than an original image, as noted in last two rows of the figure.
Notable performance improvement brought by LAN just with noise adaptation at input image suggests that noise adaptation is not just additional denoising process.

\subsection{Zero-shot denoising}
\label{sec:zeroshot}
One may argue that the performance degradation is expected with the full adaptation of a network when train noise distribution and new test noise distribution greatly differ.
Another alternative to finetuning of an network would be to train a randomly initialized network on a new noisy image from scratch via self-supervision loss functions for blind denoising.
In fact, ZS-N2N~\cite{mansour2023zero} is specifically designed for training a denoising network on a single noisy image with unknown noise.
Thus, we demonstrate such zero-shot denoising performance with DnCNN, trained via ZS-N2N and Nbr2Nbr on each image in PolyU and Nam dataset, the results of which are displayed in Table~\ref{tab:zero-shot}.
After training a network for more than 1K iterations as suggested in~\cite{mansour2023zero}, the zero-shot training yields significantly poor performance, compared to not just LAN but also SIDD-pretrained model and other alternative adaptation methods.
Despite the deviations and misalignment in noise, the results suggest the benefits of exploiting the knowledge of denoising tasks from a large training set. This is similar to domain adaptation, supporting our formulation.

\begin{table}
  \centering
    \aboverulesep=0ex
    \belowrulesep=0ex
    \resizebox{0.35\textwidth}{!}{
  \begin{tabular}{cc cc cc cc}
    \toprule
    \multirow{2}{*}{Model}& \multirow{2}{*}{Self-loss} & \multicolumn{2}{c}{LAN / Full-trainable}\\
    \cmidrule(r){3-4}
    & & Time & Memory \\
    \midrule
    \multirow{2}{*}{Restormer}& ZS-N2N & 79.88 \%  & 93.27\% \\
    \cmidrule(r){2-4}
     & Nbr2Nbr & 93.04 \% &  92.22\%  \\
    \midrule
    \multirow{2}{*}{Uformer}& \multirow{1}{*}{ZS-N2N} 
    & 74.10 \% & 73.75\% \\
    \cmidrule(r){2-4}
     & \multirow{1}{*}{Nbr2Nbr} 
    & 85.24\% & 74.21\% \\
    \bottomrule
  \end{tabular}
  }
  \vspace{-0.5em}
  \caption{The runtime and memory efficiency ratio of LAN (ours) to full-trainable based on an image size of $256\times256$.}
  \vspace{-1.5em}
  \label{tab:efficiency}
\end{table}

\subsection{Computational efficiency}
\label{sec:efficiency}
One may be concerned with the computational cost of our proposed LAN framework due to the introduction of a learnable parameter for each image pixel. However, our evaluations show that LAN is more efficient in memory and runtime including both adaptation and inference, especially for large networks like Restormer and Uformer, as displayed in Table~\ref{tab:efficiency}. This is because fewer parameters need updating compared to a 'full-trainable' adaptation method. A limitation is that the number of trainable parameters depends on the input image size, which could reduce efficiency for large images. We plan to improve this in the future. Despite this, the notable performance improvement from our framework offers a promising research direction to explore for image denoising.

\subsection{Effects of noise adaptation.}
\label{sec:effect}
\begin{figure}
    \centering
    \begin{subfigure}[b]{0.175\textwidth}
        \centering
        \includegraphics[width=\textwidth]{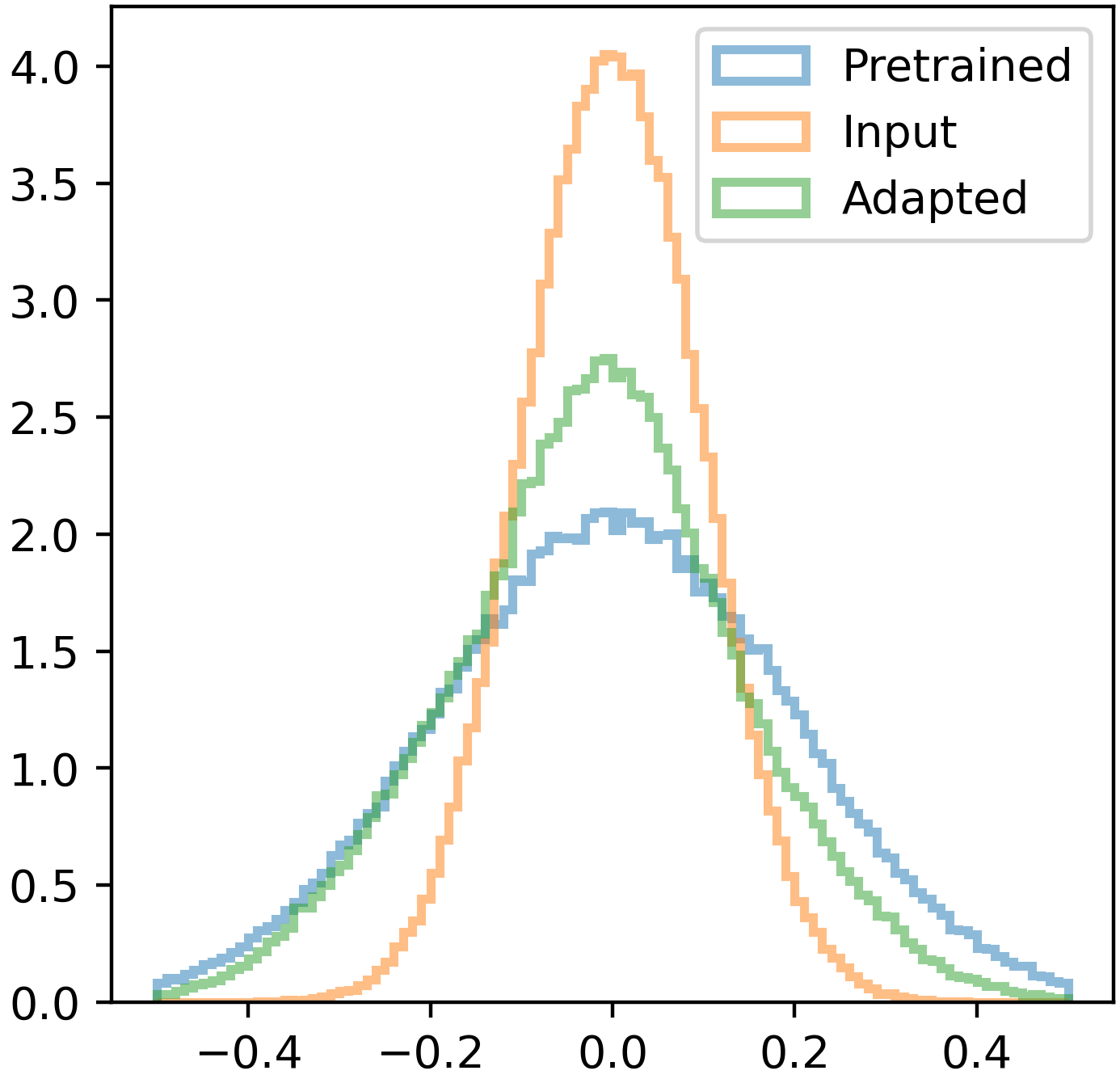}
        \caption{Adapted Gaussian noise}
        \label{gaussian_overfitting}
    \end{subfigure}
    \centering
    \hspace{0.25cm}
    \begin{subfigure}[b]{0.175\textwidth}
        \centering
        \includegraphics[width=\textwidth]{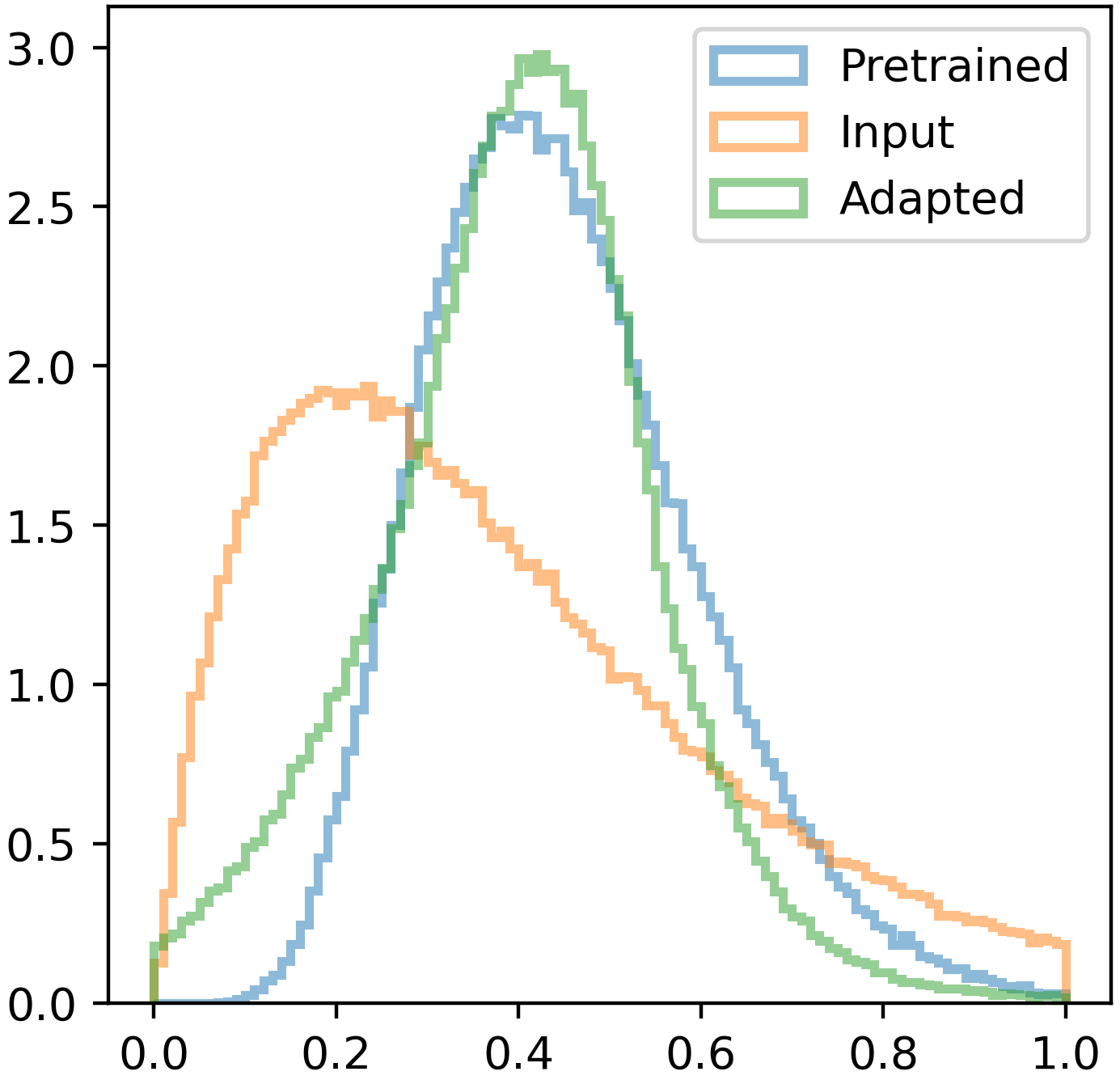}
        \caption{Adapted gamma noise}
        \label{gamma_overfitting}
    \end{subfigure}
    \centering
    \caption{
    Histogram of synthetic noise distributions.
    Adapted noise distribution (green) by LAN is shown to shift the new noise distribution (orange) to already seen noise distribution (blue).
    }
    \label{fig:noise_hist}
\end{figure}

To better illustrate the effects of noise adaptation by LAN (Equation~\ref{eq:noise_adapt}), we perform experiments with synthetic noises for clear visualization.
Specifically, Gaussian $\sigma=50$ or Gamma distribution $\theta=0.5, k=9$ noise is added to the single training data from BSDS500~\cite{bsds500:martin2001database} dataset to create synthetic-noisy-clean image pairs for pretraining the DnCNN.
Figure~\ref{fig:noise_hist} shows a histogram of noise from a synthetic training set (denoted as `Pretrained'), a new noisy image (denoted as `Input'), and an adapted noisy image (denoted as `Adapted').
The figure demonstrates that our algorithm brings a new noise closer towards a noise expected by a denoising network.
We also visualize such noisy images in Figure~\ref{fig:synthetic}, where a plain image is used for better visualization.
LAN is shown to try to adapt a new noisy image to contain similar noise used during pretraining.
In particular, we observe newly added noise on top of adapted noisy image.
The results illustrate that our noise adaptation is not just additional denoising.

\begin{table}
  \centering
  \aboverulesep=0ex
\belowrulesep=0ex
\resizebox{0.35\textwidth}{!}{
  \begin{tabular}{cc cc cc cc}
    \toprule
    Dataset& Self-loss & PSNR (dB) & SSIM \\
    \midrule
    \multirow{2}{*}{PolyU}& ZS-N2N & 31.47  & 0.875 \\
    \cmidrule(r){2-4}
     & Nbr2Nbr & 32.93 & 0.911  \\
    \midrule
    \multirow{2}{*}{Nam}& \multirow{1}{*}{ZS-N2N} 
    & 34.47 & 0.902  \\
    \cmidrule(r){2-4}
     & \multirow{1}{*}{Nbr2Nbr} 
    & 35.39 & 0.923 \\
    \bottomrule
  \end{tabular}
  }
  \caption{Zero-shot denoising performance with randomly initialized DnCNN, trained via ZS-N2N on each image in PolyU and Nam dataset.}
  \label{tab:zero-shot}
\end{table}

\section{Conclusion}
In this work, we propose a new adaptation approach to handle unseen noise for image denoising.
We focus on the fine-grained pixel-level misalignment issues between unseen noise in new noisy images and seen noise during the pretraining of a denoising network.
In contrast to standard approaches of finetuning a model, we focus on adapting an input noisy image itself.
To this end, we introduce a new denoising framework, named Learning-to-Adapt-Noise (LAN), that adds a new noisy image with a learnable offset that is trained to bring noise in a new noisy image closer to noise seen during the pretraining stage.
The experimental results solidifies the motivation and effectiveness of noise adaptation by our proposed method.
One limitation would be that the computation and resource complexities may grow with the size of input images, although LAN is more efficient in comparison to model adaptation for images of typical size $256\times256$.
Nevertheless, we believe that our work brings interesting results and research discussions, and suggests a new research direction.
\section{Acknowledgements}
This work was supported by Institute of Information \& communications Technology Planning \& Evaluation (IITP) grant funded by the Korea government (MSIT) (No.2022- 0-00156, Fundamental research on continual meta-learning for quality enhancement of casual videos and their 3D metaverse transformation), (No.2020-0-01373, Artificial Intelligence Graduate School Program (Hanyang University)).
\setcounter{section}{0}
\setcounter{figure}{0}
\setcounter{table}{0}
\setcounter{equation}{0}

\renewcommand{\thetable}{S\arabic{table}}
\renewcommand{\thesection}{S\arabic{section}}
\renewcommand{\thefigure}{S\arabic{figure}}
\renewcommand{\theequation}{S\arabic{equation}}

\newpage
\clearpage

{
    \small
    \bibliographystyle{ieeenat_fullname}
    \bibliography{main}
}

\end{document}